\documentclass{llncs}

\usepackage{amsmath, amssymb}
\DeclareMathOperator{\E}{\mathbb{E}}
\DeclareMathOperator*{\argmax}{arg\,max}
\usepackage{mathtools}
\DeclarePairedDelimiter{\norm}{\lVert}{\rVert}
\newcommand{\RomanNumeralCaps}[1]
{\MakeUppercase{\romannumeral #1}}
\usepackage{hyperref}
\usepackage{cite}
\usepackage{enumitem}

\usepackage{graphicx}
\graphicspath{{./Figs/}}

\usepackage{siunitx}
\usepackage{booktabs}
\newcommand{\bftab}{\fontseries{b}\selectfont}

\usepackage{multirow}
\usepackage{array}

\usepackage{comment}
\usepackage{caption}
\usepackage{subcaption}
\usepackage[toc,title]{appendix}

\usepackage{fancyhdr}
\begin{document}
	
\title{Tampered VAE for Improved Satellite Image Time Series Classification}

\author{Xin Cai \and Yaxin Bi \and Peter Nicholl}
\institute{School of Computing, Ulster University\\
	\email{\{cai-x, y.bi, p.nicholl\}@ulster.ac.uk}}

\maketitle	
	
\begin{abstract}
	The unprecedented availability of spatial and temporal high-resolution satellite image time series (SITS) for crop type mapping is believed to necessitate deep learning architectures to accommodate challenges arising from both dimensions. Recent state-of-the-art deep learning models have shown promising results by stacking spatial and temporal encoders. However, we present a Pyramid Time-Series Transformer (PTST) that operates solely on the temporal dimension, i.e., neglecting the spatial dimension, can produce superior results with a drastic reduction in GPU memory consumption and easy extensibility. Furthermore, we augment it to perform semi-supervised learning by proposing a classification-friendly VAE framework that introduces clustering mechanisms into latent space and can promote linear separability therein. Consequently, a few principal axes of the latent space can explain the majority of variance in raw data. Meanwhile, the VAE framework with proposed tweaks can maintain competitive classification performance as its purely discriminative counterpart when only $40\%$ of labelled data is used. We hope the proposed framework can serve as a baseline for crop classification with SITS for its modularity and simplicity.
\end{abstract}
	
\section{Introduction}

Recent advances in remote sensing has enabled monitoring the Earth surface with unprecedentedly high resolution and frequency, which is particularly helpful for the study of vegetation. For example, the recently released DENETHOR \cite{kondmann2021denethor} dataset provides harmonized, declouded, and daily Planet Fusion Surface Reflection data for crop type mapping. However, such high temporal and spatial density present significant challenges for developing deep learning models that can fully exploit the available information while maintaining the computation budgets in a reasonable limit.

Recent state-of-the-art deep learning models feature a combination of spatial and temporal encoders, e.g., the Pixel-Set spatial and the self-attention temporal encoders \cite{garnot2020satellite, garnot2020lightweight}. Despite the promising results achieved, such kind of models require prohibitively high GPU memory consumption due to the extraction of spatial descriptors from raw pixels using neural networks even with an extremely lightweight model. Seemingly, the learned spatial descriptors are superior to simple statistics, such as mean and standard deviation values of the parcel fields. However, is it truly worthwhile to sacrifice the GPU memory usage for learnable spatial descriptors for crop classification? More importantly, when it comes to SITS with dense temporal resolution, maintaining high resolution in both dimensions would easily trigger out-of-memory (OOM) even with the modern GPU architecture. We argue that the emphasis should be placed on the temporal dimension and the spatial information can be summarized with simple statistics.

Besides, current deep learning models for SITS classification contain bespoke components, which means such models are difficult to be extended with minimal modifications and hyper-parameter tuning. Therefore, it is necessary to develop highly modularized neural architectures given the widespread success of Transformers \cite{vaswani2017attention, liu2021swin}, thereby reducing the burden of researchers caused by manual architecture design from scratch. This motivates us to adapt the successful Pyramid Vision Transformer model \cite{wang2021pyramid, wang2021pvtv2} proposed for computer vision tasks to process time-series data. Surprisingly, this simple model can yield superior performance compared to the existing state-of-the-art models while only operating on the temporal dimension without the need to devise tailored spatial components. In addition, it can be easily extended to models of large capacity whenever the labelled data is abundant due to its modularity. 

More importantly, the simplicity of this baseline model enables us to divert our attention to a more significant problem for SITS classification: semi-supervised learning, given the abundance of unlabelled satellite data and the laborious annotation process. Variational Auto-Encoder \cite{DBLP:journals/corr/KingmaW13, kingma2019introduction} (VAE) is a generalized and scalable probabilistic approach for semi-supervised learning \cite{kingma2014semi}, where the probabilistic encoder can be regarded as a discriminative classifier and the probabilistic decoder can be employed to match the generated data distribution to the empirical one. However, most of work has focused on the fully unsupervised setting or learning latent disentangled factors of variation \cite{DBLP:conf/iclr/HigginsMPBGBML17, DBLP:conf/nips/Dupont18}. Regarding the use of VAEs for semi-supervised learning \cite{DBLP:conf/conll/ChengXWCHCH19}, the optimization objective still sticks to the one proposed in the seminal work \cite{kingma2014semi} where scalable stochastic variational inference \cite{paisley2012variational} (SVI) has been first introduced for semi-supervised classification. We argue that the vanilla VAE objective is not friendly to achieve classification accuracy as high as purely discriminative classifiers. Therefore, we propose an augmented variant of the vanilla VAE objective which is beneficial for establishing a more linearly separable latent space and has experimentally proven to be more effective both in the fully- and semi- supervised setting for crop classification. Additionally, we provide theoretical analysis to justify the proposed modifications.

\section{Related Work}

Crop type mapping using SITS has received considerable attention because of its huge potential in crop phenology study, crop yield prediction, facilitating informed decisions on subsidy grants, and food security estimation. With the increasing availability of satellite data, significant research efforts have been devoted to developing automated tools for efficient analysis. Traditionally, manually crafted features, such as vegetation indices \cite{hao2015feature}, combined with machine learning algorithms, such as Random Forest classifier \cite{valero2016production} and Support Vector Machine \cite{devadas2012support}, have been the norm for crop classification. Recently, deep learning-based models have dominated the research field by adapting successful neural architectures developed in other domains. The existing approaches generally fall into two categories, pure time-series models and the combination of spatial and temporal encoders. The former one is focused on capturing temporal dynamics by using sequential neural models, such as Recurrent Neural Networks \cite{russwurm2018multi, russwurm2017temporal}, Temporal Convolution Neural Networks \cite{ismail2020inceptiontime, pelletier2019temporal}, and self-attention models \cite{russwurm2020self}. The latter one argues that extraction of spatial features using neural networks is superior to simple statistics, mainly following the paradigm of obtaining spatial embeddings with spatial encoders and then processing these embeddings sequentially with temporal encoders. Along this line of research, recent work \cite{garnot2020satellite} has pointed out that convolutions are not well-suited for extracting spatial features from satellite data for crop classification due to the highly irregular boundaries of parcel fields and limited texture patterns available. Consequently, the researchers have proposed to use Pixel-Set Encoder \cite{garnot2020satellite} to obtain learnable statistical descriptors, which is inspired by advances in 3D point cloud processing \cite{qi2017pointnet}. The resulted spatial embeddings are then processed by temporal encoders using self-attention mechanisms \cite{vaswani2017attention}, leveraging its power for capturing long-range dependencies. Furthermore, the lightweight temporal attention module has been proposed in \cite{garnot2020lightweight} by dividing feature vectors into different groups with specialized attention weights calculated for each group. This module can be considered as a strengthened extension of group convolution, which is the core computing unit of ResNeXt \cite{xie2017aggregated}.

\section{Proposed Method}

\subsection{Preliminaries}

Variational Auto-Encoders provide a framework for approximate posterior inference and log-likelihood estimation by using the reparameterization trick \cite{kingma2019introduction, rezende2014stochastic}, which enables the objective can be efficiently optimized through Stochastic Gradient Descent (SGD). Generally, the objective of VAEs can be formulated as in Eq. \eqref{eq:3-1}:

\begin{align}
	\log{p_{\theta}\left(\mathbf{x}\right)} &= \log{\E_{q_{\phi}\left(\mathbf{z} | \mathbf{x}\right)}}\left[\frac{p_{\theta}\left(\mathbf{x}, \mathbf{z}\right)}{q_{\phi}\left(\mathbf{z} | \mathbf{x} \right)}\right] \geq \E_{q_{\phi}\left(\mathbf{z} | \mathbf{x}\right)}\left[\log{\frac{p_{\theta}\left(\mathbf{x}, \mathbf{z}\right)}{q_{\phi}\left(\mathbf{z} | \mathbf{x}\right)}}\right] \nonumber \\	
	\mathcal{L}_{\theta, \phi}\left(\mathbf{x}\right) &= \E_{q_{\phi}\left(\mathbf{z} | \mathbf{x}\right)}\left[\log{p_{\theta}\left(\mathbf{x}, \mathbf{z}\right)} - \log{q_{\phi}\left(\mathbf{z} | \mathbf{x}\right)} \right] \nonumber \\
	&= \E_{q_{\phi}\left(\mathbf{z} | \mathbf{x}\right)}\left[\log{p_{\theta}\left(\mathbf{x} | \mathbf{z}\right)}\right] - D_{\mathrm{KL}}\left(q_{\phi}\left(\mathbf{z} | \mathbf{x}\right) \Vert p_{\theta}\left(\mathbf{z}\right) \right) \label{eq:3-1}
\end{align}

\noindent where $\mathcal{L}_{\theta, \phi}\left(\mathbf{x}\right)$ is referred to as individual-datapoint evidence lower bound (ELBO), which consists of the probabilistic encoder $q_{\phi}\left(\mathbf{z} | \mathbf{x}\right)$, the probabilistic decoder $p_{\theta}\left(\mathbf{x} | \mathbf{z}\right)$, and a regularization term between the approximate posterior and prior distributions.

One of the obstacles caused by incorporating stochastic nodes, such as $\mathbf{z}$, into computation graphs is blocking the path of gradients for back-propagation, which can be neatly sidestepped by reparameterizing latent variables $\mathbf{z}$ as a deterministic and differentiable function of an externalized randomness source:

\begin{align}
	\boldsymbol{\epsilon} &\sim p\left(\boldsymbol{\epsilon}\right) \nonumber\\
	\mathbf{z} &= \mathrm{Encoder}_{\phi}\left(\mathbf{x}, \boldsymbol\epsilon \right) \nonumber\\
	\mathcal{L}_{\theta, \phi}\left(\mathbf{x}\right) &= \E_{p\left(\boldsymbol{\epsilon}\right)}\left[\log{p_{\theta}\left(\mathbf{x}, \mathbf{z} \right)} - \log{q_{\phi}\left(\mathbf{z}|\mathbf{x} \right)} \right] \nonumber \\
	 &\approx \frac{1}{n} \sum_{i=1}^n \log{p_{\theta}\left(\mathbf{x}, \mathbf{z}_i \right)} - \log{q_{\phi}\left(\mathbf{z}_i | \mathbf{x} \right)} = \tilde{\mathcal{L}}_{\theta, \phi}\left(\mathbf{x} \right) \label{eq:3-2}
\end{align}
  
Therefore, the Eq. \eqref{eq:3-2} implies that the reparameterization trick enables the gradients of the ELBO $\nabla \mathcal{L}_{\theta, \phi}\left(\mathbf{x}\right)$ w.r.t both generative model and variational parameters $\theta$ and $\phi$ to be calculated with Monte Carlo estimates $\nabla \tilde{\mathcal{L}}_{\theta, \phi}\left(\mathbf{x}\right)$, which is an unbiased estimator of the exact gradients of the ELBO. 

Continuous latent random variables can be easily dealt with using the reparameterization trick either by explicitly specifying a parameterized posterior distribution, such as a spherical Gaussian distribution, or implicitly learned from the data by ensuring that transformations of random variables are not only differentiable but invertible, which is referred to as the Normalizing Flow technique\cite{DBLP:conf/iclr/DinhSB17, kingma2016improved, kingma2018glow}. Regarding discrete random variables, continuous relaxations need to be employed to ensure differentiability. Specifically, latent categorical random variables can be approximated by the continuous Concrete/Gumbel-Softmax random variables, independently discovered in the work \cite{DBLP:conf/iclr/MaddisonMT17, DBLP:conf/iclr/JangGP17}, which are defined as follows:

\begin{align}
	Y_k = \frac{\exp{\left(\left(\log{\alpha_k} + G_k\right) / \lambda\right)}}{\sum_{i=1}^K \exp{\left(\left(\log{\alpha_i} + G_i\right) / \lambda\right)}} \label{eq:3-3}
\end{align} 

\noindent where $Y \in \Delta^{K-1} = \left\{y \in \mathbb{R}^K \mid y_k \in \left[0, 1\right], \sum_{i=1}^K y_i = 1 \right\}$ with the temperature $\lambda \in \left(0, \infty\right)$, (unnormalized)parameters of the categorical distribution $\alpha_k \in \left(0, \infty\right)$, and i.i.d. gumbel noise $G_k \sim \mathrm{Gumbel}\left(0, 1\right)$. The property of Concrete random variables $	\mathbb{P}\left(\lim_{\lambda\to 0} Y_k=1\right)=\frac{\alpha_k}{\sum_{i=1}^{K}\alpha_i}$ guarantees that when the temperature $\lambda$ approaches $0$, the samples of the Concrete distribution can well approximate those drawn from $\argmax \left(\log\alpha_k + G_k\right) \sim \mathrm{Cat}\left(\boldsymbol\pi \mid \pi_i = \frac{\alpha_i}{\sum_{i=1}^{K}\alpha_i} \right)$, which is the well-known Gumbel-Max trick \cite{maddison2014sampling}.

\subsection{Tampered VAE Objective}
\label{sec:3-2}

A simple extension of the Eq. \eqref{eq:3-1} by explicitly separating discrete latent variables from continuous ones lends VAEs to utilize both labelled and unlabelled data to perform semi-supervised classification, which was originally proposed in the work \cite{kingma2014semi} and then adapted by replacing the costly marginalization over categorical variables by the reparameterizable Concrete random variables \cite{DBLP:conf/iclr/JangGP17}, thus allowing both efficient forward and backward computation.

Generally, the continuous and discrete latent variables are assumed to be conditionally independent \cite{DBLP:conf/nips/Dupont18}. We argue, however, that explicitly considering the dependency between the continuous and categorical random variables $\mathbf{z}$ and $y$ can lead to a more classification-friendly VAE. Following the assumption, the objective can be formulated as follows:  

\begin{align}
	\mathcal{L}_{\theta, \phi}\left(\mathbf{x}\right) =& 
	\E_{q_{\phi}\left(y, \mathbf{z} | \mathbf{x}\right)}\left[\log{p_{\theta}\left(\mathbf{x}, y, \mathbf{z}\right)} - \log{q_{\phi}\left(y, \mathbf{z} | \mathbf{x}\right)}\right] \nonumber \\
	=& \E_{q_{\phi}\left(y | \mathbf{x} \right)}\E_{q_{\phi}\left(\mathbf{z} | \mathbf{x}, y \right)} \left[\log p_{\theta}\left(\mathbf{x} | \mathbf{z}, y \right) \right] \nonumber \\  &- \E_{q_{\phi}\left(y | \mathbf{x}\right)} \left[D_{\mathrm{KL}}\left(q_{\phi}\left(\mathbf{z} | \mathbf{x}, y \right) \Vert p_{\theta}\left(\mathbf{z} | y \right) \right) \right] \nonumber \\
	&- D_{\mathrm{KL}}\left(q_{\phi}\left(y | \mathbf{x}\right) \Vert p_{\theta}\left(y\right) \right) \label{eq:3-4}
\end{align}

The Eq. \eqref{eq:3-4} naturally implies a mixture distribution over the latent space, which bears striking similarity to the Gaussian Mixture Model (GMM) \cite{bishop2006pattern}, provided that $p_{\theta}\left(\mathbf{z} | y\right)$ is parameterized by multivariate Gaussians and  $p_{\theta}\left(y\right) \sim \mathrm{Cat}\left(\boldsymbol{\pi}\right)$. For simplicity, in this paper, we adopted the assumption that the continuous latent random variable can be characterized by spherical Gaussians. Rather than enforcing the disentanglement between the continuous and categorical latent variables $\mathbf{z}$ and $y$ \cite{DBLP:conf/nips/Dupont18}, which are commonly interpreted as ``style" and class-specific latent representations, introducing the dependency between $\mathbf{z}$ and $y$ imposes the Mixture of Gaussians (MoG) prior over latent variables which has been shown to be effective for reducing useless latent dimensions of $\mathbf{z}$, therefore alleviating the posterior collapse problem of VAEs to a certain extent \cite{tomczak2018vae}. Following this line of reasoning, we argue that enforcing the cross-entropy loss on latent variable $\mathbf{z}$ can reinforce this effect when labels are available. 

Despite the theoretically sound formulation, we found that the second term in the Eq. \eqref{eq:3-4} inherently would hamper the objective of assigning continuous latent features $\mathbf{z}$ to their respective classes, i.e., making $\mathbf{z}$ discriminative or separable. It is tempting to consider that the mode-search behavior of the reverse KL divergence $D_{\mathrm{KL}}\left(q_{\phi}\left(\mathbf{z} | \mathbf{x}, y \right) \Vert p_{\theta}\left(\mathbf{z} | y \right) \right)$ is beneficial for matching $q_{\phi}\left(\mathbf{z} | \mathbf{x}, y \right)$ to its ground-truth component. However, when the recognition model $q_{\phi}\left(y|\mathbf{x}\right)$ is unstable and weak, especially during the early stage of training, the regularization term is prone to drive the approximate posterior to a trivial prior as verified in our experiments, when setting a learnable MoG prior, class centers would become indistinguishable, which is consistent with findings in other work \cite{DBLP:conf/conll/BowmanVVDJB16, DBLP:journals/corr/SonderbyRMSW16, DBLP:conf/naacl/FuLLGCC19} where they proposed to use scheduling schemes to mitigate this negative effect. Besides, following the analysis in $\beta$-VAE \cite{DBLP:conf/iclr/HigginsMPBGBML17}, the minimization of the KL term $D_{\mathrm{KL}}\left(q_{\phi}\left(\mathbf{z} | \mathbf{x}, y \right) \Vert p_{\theta}\left(\mathbf{z} | y \right) \right)$ would force conditional independence between continuous and discrete latent variables $\mathbf{z}$ and $y$, which we argue would produce detrimental effects on classification performance. As class information is ubiquitous, even though it has not been explicitly treated in the common VAE objective, the Eq. \eqref{eq:3-4} can partially provide a theoretical explanation for the posterior collapse problem that plagues VAEs. 

Based on the previous analysis, we propose to replace the second term in Eq. \eqref{eq:3-4} with the cosine similarity measure, which can evaluate the distance not only between latent $\mathbf{z}$ and its ground-truth class center but also the centers of the negative classes. Introducing the cosine similarity is beneficial for establishing a latent space with features having reduced intra-class distances and increased inter-class distances, which is the core objective for many clustering methods and serves the similar purpose of imposing a cross-entropy loss on latent $\mathbf{z}$ but is suitable when labels are not available. Furthermore, the optimization of the cosine similarity term can inject positive/negative class information into latent dimensions of the variable $\mathbf{z}$, which can be regarded as an upgraded alternative of the free-bits VAE objective \cite{kingma2016improved}.

Combining those modifications proposed previously, we have reached the ultimate objective, which is one of the core contributions of this paper and formulated as follows:

\begin{align}
	\mathcal{J}_{\theta, \phi}\left(\mathbf{x}\right) =& 
	 -\E_{q_{\phi}\left(y | \mathbf{x} \right)}\E_{q_{\phi}\left(\mathbf{z} | \mathbf{x}, y \right)} \left[\log p_{\theta}\left(\mathbf{x} | \mathbf{z}, y \right) \right] + \E_{q_{\phi}\left(y | \mathbf{x}\right)} \left[\mathrm{Cos}\left(\mathbf{z}, \mathbf{c}_y \right) \right] \nonumber \\ 
	 &+ \gamma_1 \E_{p\left(y| \mathbf{x}\right)}\left[-\log q_{\phi}\left(y | \mathbf{x}\right) - \log q_{\eta}\left(y | \mathbf{z} \right) \right] \nonumber \\
	 &+ \gamma_2 D_{\mathrm{KL}}\left(q_{\eta}\left(y|\mathbf{z}\right) \Vert q_{\phi}\left(y|\mathbf{x}\right) \right) + D_{\mathrm{KL}}\left(q_{\phi}\left(y | \mathbf{x}\right) \Vert p_{\theta}\left(y\right) \right) \label{eq:3-5} \\ 
	 \mathrm{Cos}\left(\mathbf{z}, \mathbf{c}_y\right)&=
	 \begin{cases}
	 	1 - \frac{\mathbf{z}^{\intercal}\mathbf{c}_y}{\norm{\mathbf{z}}\norm{\mathbf{c}_y}} & \text{if $y=1$}\\
	 	\max\left(0, \frac{\mathbf{z}^{\intercal}\mathbf{c}_y}{\norm{\mathbf{z}}\norm{\mathbf{c}_y}} - margin\right) & \text{if $y=-1$}\\
	 \end{cases} \nonumber
\end{align}

\noindent where $\gamma_1$ and $\gamma_2$ are two hyper-parameters to balance the respective contribution towards the total loss, $q_{\eta}\left(y|\mathbf{z}\right)$ denotes an auxiliary classifier for latent features, and the term $D_{\mathrm{KL}}\left(q_{\eta}\left(y|\mathbf{z}\right) \Vert q_{\phi}\left(y|\mathbf{x}\right) \right)$ is for refining the crude recognition model by conveying extra information from the predicted distribution based on $\mathbf{z}$ as we presume latent features are easier to be classified given the separability promoted by the strong regularization. Substituting the term $\E_{q_{\phi}\left(y | \mathbf{x}\right)} \left[D_{\mathrm{KL}}\left(q_{\phi}\left(\mathbf{z} | \mathbf{x}, y \right) \Vert p_{\theta}\left(\mathbf{z} | y \right) \right) \right]$ with our proposed one $\E_{q_{\phi}\left(y | \mathbf{x}\right)} \left[\mathrm{Cos}\left(\mathbf{z}, \mathbf{c}_y \right) \right]$ is beneficial for counteracting the undesirable effects of over regularization such that the notorious KL-vanishing problem in the vanilla VAE has been turned into one of the desired optimization objectives, which will be proved in Section \ref{sec:4}. Besides, we chose to incorporate the term $D_{\mathrm{KL}}\left(q_{\phi}\left(y | \mathbf{x}\right) \Vert p_{\theta}\left(y\right) \right)$ for completeness but found it has no significant contribution towards improving the classification accuracy in our experiments even though it can be considered a type of label smoothing. Our proposed objective \eqref{eq:3-5} can be regarded as an augmented variant of the vanilla VAE objective, which is particularly refined for building a strong classifier. More importantly, the objective \eqref{eq:3-5} can be easily adapted for semi-supervised learning by substituting the approximate posterior $q_{\phi}\left(y|\mathbf{x}\right)$ with the true posterior $p\left(y|\mathbf{x}\right)$ where labels are accessible and removing unlabelled data from the calculation of the cross-entropy loss, which is also the reason why we do not explicitly differentiate $y$ in Eq. \eqref{eq:3-5} for predicted or ground-truth values.

\subsection{Model Instantiation}

Based on the analysis in Sec. \ref{sec:3-2}, we propose to implement the probabilistic encoder with a Pyramid Transformer architecture tailored for time-series data and the probabilistic decoder with a simple Transformer model without upsampling operations. The neural architecture design is heavily inspired by recent advances in Transformer models \cite{wang2021pyramid, wang2021pvtv2, chu2021conditional, nguyen2019transformers, chen-etal-2018-best} to reduce the potentially prohibitive cost of verifying building components and hyperparameter tuning while maintaining the model as simple as possible, which is beneficial for demonstrating the effectiveness of the proposed modifications to the vanilla VAE objective.

The probabilistic encoder is adapted from the recently proposed Pyramid Vision Transformer (PVT) \cite{wang2021pyramid, wang2021pvtv2}, which is composed of the overlapping patch embedding, conditional position encoding (CPE), multi-head self-attention (MHSA), and feed-forward (FFD) layers combined as the basic building unit, as illustrated in Fig. \ref{fig: a1-1} in Appendix. The patch embedding layer is used for tokenizing sequential data and performing progressive shrinkage of the resolution of feature maps with strides greater than $1$. The CPE proposed in CPVT \cite{chu2021conditional} is essentially depth-wise convolutions, which is similar to methods proposed in \cite{li2021localvit, yuan2021tokens}, to introduce locality mechanisms into Transformer models to complement the self-attention operation which is adept at capturing long-range dependencies between tokens. Besides, the CPE is more flexible than absolute position encoding \cite{vaswani2017attention} when processing variable-sized inputs. The pyramid structure has been widely-adopted for recognition tasks in the vision domain \cite{he2016deep, liu2021swin, wang2021pyramid}, which has proven to be effective for extracting hierarchical and high-level semantic features and also can significantly reduce computation consumption with increased dimensions of feature maps. However, the simple adaptation of PVT cause optimization difficulties during the training. Therefore, we further adopted modifications proposed in \cite{nguyen2019transformers, chen-etal-2018-best} to ease the difficulty of training Transformers, especially when trained from scratch and on moderately-sized datasets, which are the PreNorm strategy and the Scale $\ell_2$ Normalization layer, as described in Eq. \eqref{eq:3-6}:

\begin{align}
	&\mathrm{ScaleNorm}\left(\mathbf{h}, g\right) = g\frac{\mathbf{h}}{\norm{\mathbf{h}}} \nonumber\\
	&\mathbf{h}_{\ell+1} = \mathbf{h}_{\ell} + \mathrm{F}_{\ell}\left(\mathrm{ScaleNorm}\left(\mathbf{h}_{\ell}, g\right)\right) \label{eq:3-6}
\end{align}

\noindent where $g$ is a learnable scalar, $\mathbf{h}$ denotes hidden representations, and $\mathrm{F}_{\ell}$ represents any transformation layer.

With all these modifications combined, we have obtained our baseline model: Pyramid Time-Series Transformer (PTST), which is also the recognition component in its VAE extension. It should be noted that this baseline model is highly modularized, which means the model capacity can be easily controlled by tuning the following hyperparameters \footnote{Please refer to Tab. \ref{tab:a1-1} in Appendix for their specific settings.}: 
\begin{itemize}[label=$\bullet$]
	\item $P_i$: the kernel size of the patch embedding layer of the stage $i$
	\item $S_i$: the stride of patch embedding layer of the stage $i$
	\item $C_i$: the channel number of the output of the stage $i$
	\item $L_i$: the number of encoder layers in the stage $i$
	\item $H_i$: the number of heads in the MHSA layer in the stage $i$
	\item $E_i$: the expansion rate of the FFD layer in the stage $i$
\end{itemize}

The generative component is composed of stacking multiple basic building blocks in Fig. \ref{fig: a1-1} and takes as input the learnable position embeddings with its sequence length equal to that of the input \footnote{When the sequence length is set to be less than that of the input for computation considerations, the upsampling layer is appended at last.} and continuous latent variable $\mathbf{z}$ replicated along the temporal dimension.

\section{Experiment}
\label{sec:4}

\subsection{Datasets \& Evaluation Metrics}

We evaluate our proposed models, both the PTST and its VAE extension TV-PTST, on two public datasets, which are part of the recent AI4EO Food Security Challenge\footnote{\url{https://platform.ai4eo.eu/}}. The challenge covers two areas of interest (AOI), in Germany and South Africa, with high-quality cadastral data on field boundaries and crop types as ground truth input. The notebook \footnote{\url{https://github.com/AI4EO/tum-planet-radearth-ai4food-challenge/blob/main/notebook/starter-pack.ipynb}} officially provided by the organizers showcases how to access and preprocess satellite data for crop type classification. It should be noted that the data collected in Germany is identical to the one released in the recent paper \cite{kondmann2021denethor} called DENETHOR.

Although the data is available in three different modalities: Sentinel-1 radar, Sentinel-2 optical, and Planet Fusion data, we only focus on the last data modality, which consists of 4 spectral bands (RGB + Near-infrared (NIR)) with a 3m spatial resolution and daily time interval \footnote{In our experiments, we only used 5-day Plant Fusion data, which means the data is median filtered with a size of five days and the same stride to reduce the temporal dimension to save the storage space and data loading time significantly.}. The Planet Fusion time series is an analysis-ready data (ARD), preprocessing level 3 product with clouds and shadows removed and the potential gaps caused by clouds filled with different points in time \cite{kondmann2021denethor}. Consequently, we found that results obtained on such type of data modality are more robust to the geographical and time shift. The train and test data in Germany are collected not only in different geographical locations but in two different years 2018 and 2019, therefore presenting the challenge of out-of-year generalization. Due to the limited number of training samples in Germany (2534) and the unavailability of the official validation dataset, we decided to use the data in South Africa for ablation study. Additionally, the train and test tiles of South Africa are  within the same year but geographically separated, making it suitable for validation without the risk of overfitting the training dataset \footnote{The commonly used k-fold cross-validation is not suitable as the identical geographical and time conditions would easily cause overfitting. Therefore, we adopted a strategy by following common practice in computer vision community.}.

Given that both datasets are highly imbalanced, we decided to report Precision (Macro), Recall (Macro), and F1 score (Macro) in addition to  Overall Accuracy (O.A.) for better reflecting the performance of classifiers \footnote{To avoid clutter, Macro will be omitted hereafter.}.

\subsection{Implementation Details}

For ablation studies, we train and validate our models on South Africa dataset, which consists of \num{4141} train and \num{2417} test samples, respectively.We use the Adam \cite{DBLP:journals/corr/KingmaB14} optimizer with an initial learning rate set to \num{1e-4} and adjusted by a linearly decay learning rate scheduler. The batch size of \num{64} and a weight decay of \num{4e-5} is employed. $\gamma_1$ is set to \num{0.1} and $\gamma_2$ is adjusted according to a cosine scheduler from \num{0.0} to \num{1.0} unless specified otherwise. All models are trained from scratch on 4 V100 GPUs for \num{40} epochs. It should be noted that these training hyper-parameters have been specifically searched for the PTST and kept fixed for TV-PTST to ensure a fair comparison as possible as we can. In addition, we followed the hyper-parameter setting for PSE+L-TAE in \cite{kondmann2021denethor} and chose a different initial learning rate \num{2e-4} to achieve better results with other training hyper-parameters unchanged. The details of  architecture design of PTST and TV-PTST are shown in Tab. \ref{tab:a1-1} and Tab. \ref{tab:a1-2} in Appendix. 

For obtaining results on DENETHOR \cite{kondmann2021denethor}, we researched training hyper-parameters and had model hyper-parameters remained the same as those for ablation studies. The initial learning rate of \num{2e-4} and \num{60} epochs are employed for training.

\subsection{Ablation Studies}

We evaluate the independent contribution of various proposed components in this section on South Africa dataset with PSE+L-TAE as the baseline model.

Firstly, we evaluate the design of PTST, which will be used as the recognition model for its VAE extension. As seen in Tab. \ref{tab:4-3-1}, the increases in O.A. ($+1.73$), Precision($+0.37$), Recall ($+4.14$), and F1 ($+2.35$) confirm the validity of replacing LayerNorm with ScaleNorm, which is consistent with observed improvements in NLP tasks\cite{nguyen2019transformers}. Notably, the pyramid shrinkage strategy improves the performance in terms of all metrics significantly compared to maintaining the temporal resolution throughout all stages (FullRes w/ SN). Also, it can be seen that including standard deviation values as model inputs in addition to mean values leads to an increase of $1.32$ in F1 score, suggesting the benefits of using more statistics to summarize parcel fields. Finally, the PTST achieves increases of $+1.69$, $+2.28$, $+3.97$, and $+3.52$ in O.A., Precision, Recall, and F1 score respectively, demonstrating the superiority of the proposed PTST for crop classification compared to the state-of-the-art PSE+L-TAE \footnote{The hyper-parameters for PST+L-TAE are specifically tuned for attaining high accuracy as possible as we can.}.

\begin{table}[!htb]
	\renewcommand{\arraystretch}{1.1}
	\setlength{\tabcolsep}{3.0pt}
	\centering
	\caption{Ablation study of our design choices of PTST. SN stands for ScaleNorm. Except for the third entry in PTST, other implementations all adopted the pyramid shrinkage strategy.}
	\label{tab:4-3-1}
	\begin{tabular}{cc|c|c|c|c}
		\toprule
	    & & O.A. $\%$ & Precision $\%$ & Recall $\%$ & F1 Score $\%$  \\ 
		\midrule
		\multicolumn{2}{c|}{PSE+L-TAE} & 73.69 & 66.40 & 59.56 & 61.79 \\ \cline{1-2}
		\multirow{4}{*}{PTST}&LayerNorm &  73.65 & 68.31 &  59.39 & 62.96 \\
		&ScaleNorm & \bftab{75.38}  & \bftab{68.68} & \bftab{63.53} & \bftab{65.31} \\
		&FullRes w/ SN & 71.78 & 65.85 & 55.78 & 56.83 \\
		&Input w/o std & 74.02 & 68.21 & 62.01 & 63.99 \\
		\bottomrule
	\end{tabular}
\end{table}

With the powerful PTST as the recognition model, we further verify that the proposed modifications to the vanilla VAE objective can lead to a more classification-friendly probabilistic model. The first column (\RomanNumeralCaps{1}) of Tab. \ref{tab:4-3-2} shows slightly better results in terms of all metrics caused by introducing probabilistic components into PTST with ground-truth information injected to the cosine loss and learnable class centers, compared to the purely discriminative model PTST. Then, we impose cross-entropy loss on latent code $\mathbf{z}$, as shown in the second column (\RomanNumeralCaps{2}) of Tab. \ref{tab:4-3-2}, which causes a performance decrease. We hypothesize such undesirable effect is caused by the inconsistency between predictions made by the recognition model and the auxiliary classifier on $\mathbf{z}$. Therefore, we additionally incorporate the $D_{\mathrm{KL}}\left(y_z, y_x\right)$, denoting the term $D_{\mathrm{KL}}\left(q_{\eta}\left(y|\mathbf{z}\right) \Vert q_{\phi}\left(y|\mathbf{x}\right) \right)$ in Eq. \eqref{eq:3-5}, into the training objective, which leads to the full realization of Eq. \eqref{eq:3-5}, as shown in the last column (\RomanNumeralCaps{7}) of Tab. \ref{tab:4-3-2}. The results of the full realization of our proposed objective show significant improvements in all metrics compared to the one implemented according to the first column (\RomanNumeralCaps{1}) and thus verify our previous assumption concerning the inconsistency. We further examine a case where $D_{\mathrm{KL}}\left(y_z, y_x\right)$ is multiplied by a constant scalar $1.0$ rather than the adjustable one that varies according to a monotonically increasing scheduling scheme, as shown in the third column (\RomanNumeralCaps{3}) of Tab. \ref{tab:4-3-2},  where noticeable drops in all metrics can be observed compared to column \RomanNumeralCaps{7}, which additionally suggests that the inconsistency between two different classifiers is dynamic. Specifically, latent code $\mathbf{z}$ is not as distinguishable as the features extracted by the recognition component in the initial period of training but would become easier to be classified towards the end because of the separability in the latent space promoted by the cosine loss. Furthermore, we substitute the ground-truth labels in the cosine loss with ones predicted by the recognition model, as shown in the fourth column (\RomanNumeralCaps{4}) of Tab. \ref{tab:4-3-2}. As expected, the use of ground-truth can facilitate the separability of the latent code, which is key for preventing the other two terms $\mathrm{CE\_loss}\left(\mathbf{z}\right)$ and $D_{\mathrm{KL}}\left(y_z, y_x\right)$ to produce counteractive effects on the classification performance as the scores in the fourth column (\RomanNumeralCaps{4}) are lower than those in the first column (\RomanNumeralCaps{1}) where these two terms are excluded from the training objective. In the fifth column (\RomanNumeralCaps{5}) of Tab. \ref{tab:4-3-2}, we tested the validity of making class centers learnable or an alternative in which orthonormal class centers are kept fixed since the initialization to enforce separability. The results of having fixed orthonormal class centers are even slightly lower than those obtained by making class centers learnable, suggesting that class centers can be made sufficiently separated during training even without explicitly imposing such restrictions. Lastly, we report the results of adding $D_{\mathrm{KL}}\left(y_{cos}, y_x\right)$ to the training objective , given the separability of latent code $\mathbf{z}$, by adding a softmax layer that takes as input cosine similarity scores calculated between $\mathbf{z}$ and class centers. But we have observed significant performance decreases compared to column \RomanNumeralCaps{7}, suggesting the insufficiency of linearity of cosine similarity to make accurate predictions. By conducting an extensive ablation study for various components proposed in Eq. \eqref{eq:3-5}, we can safely conclude that our proposed objective works in synergy to benefit classification performance.


\begin{table}[!htb]
	\renewcommand{\arraystretch}{1.1}
	\setlength{\tabcolsep}{3.0pt}
	\centering
	\caption{Ablation study of various components proposed in Eq. \eqref{eq:3-5} for making VAEs more classification-friendly. GT: ground-truth labels.}
	\label{tab:4-3-2}
	\begin{tabular}{c|ccccccc}
		\toprule
		& \RomanNumeralCaps{1} & \RomanNumeralCaps{2} & \RomanNumeralCaps{3}  & \RomanNumeralCaps{4}  & \RomanNumeralCaps{5}  & \RomanNumeralCaps{6}  &  \RomanNumeralCaps{7} \\
		\midrule
		$\mathrm{CE\_loss}\left(\mathbf{z}\right)$ &  & \checkmark & \checkmark & \checkmark & \checkmark & \checkmark & \checkmark \\
		$D_{\mathrm{KL}}\left(y_z, y_x\right)$ const. & & & \checkmark & & & & \\
		$D_{\mathrm{KL}}\left(y_z, y_x\right)$ sched. & & & & \checkmark & \checkmark & \checkmark & \checkmark \\
		$\mathrm{Cos}\left(\mathbf{z}, \mathbf{c}_y \right)$ w/ GT & \checkmark & \checkmark & \checkmark & & \checkmark & \checkmark & \checkmark \\
		learnable cls centers & \checkmark & \checkmark & \checkmark & \checkmark & & \checkmark & \checkmark \\
		$D_{\mathrm{KL}}\left(y_{cos}, y_x\right)$ sched. & & & & & & \checkmark &\\
		\hline
		O.A. $\%$ & 75.47  & 74.68 & 75.30 & 75.47 & 76.46 & 75.76 & \bftab{76.79} \\
		Precision $\%$ & 69.64 & 67.45 & 70.81 & 69.21 & \bftab{73.22} & 70.58 & 72.41\\
		Recall $\%$ & 64.11 & \bftab{65.74} & 62.35 & 63.34 & 64.76 & 63.85 & 65.32\\
		F1 Score $\%$ & 66.08 & 65.89 & 65.31 & 65.76 & 67.96 & 66.64 & \bftab{68.33}\\
		\bottomrule
	\end{tabular}
\end{table}

\begin{table}[!htb]
	\renewcommand{\arraystretch}{1.1}
	\setlength{\tabcolsep}{3.0pt}
	\centering
	\caption{Comparison between our proposed $\mathrm{Cos}\left(\mathbf{z}, \mathbf{c}_y \right)$ and the original $D_{\mathrm{KL}}\left(q_{\phi}\left(\mathbf{z} | \mathbf{x}, y \right) \Vert p_{\theta}\left(\mathbf{z} | y \right) \right)$. Y: predictions from the recognition model. Z: predictions from the auxiliary classifier on latent code $\mathbf{z}$. Cos: predictions from cosine similarity scores between $\mathbf{z}$ and class centers.}
	\label{tab:4-3-3}
	\begin{tabular}{cc|c|cc|cc|cc}
		\toprule
		&&class&\multicolumn{2}{c|}{Y} & \multicolumn{2}{c|}{Z} & \multicolumn{2}{c}{Cos} \\
		&&centers & O.A.$\%$ & F1$\%$ & O.A.$\%$ & F1$\%$ & O.A.$\%$ & F1$\%$ \\
		\midrule
		\multirow{4}{*}{Part} & \multirow{2}{*}{$D_{\mathrm{KL}}$} & fixed & 75.67 & 67.21 & -- & -- & 75.59 & 66.34  \\
		& & learn & 75.09 & 66.23 & -- & -- & 5.71 & 2.16 \\
		& \multirow{2}{*}{$\mathrm{Cos}$} & fixed & 75.22 & 65.70 & -- & -- & 75.71 & 66.39 \\
		& & learn & 75.47 & 66.08 & -- & -- & 74.76 & 64.43 \\
		\hline
		\multirow{4}{*}{Full} & \multirow{2}{*}{$D_{\mathrm{KL}}$} & fixed & 69.88 & 52.47 & 68.14 & 43.11 & 74.89 & 64.08  \\
		& & learn & 71.20 & 56.35 & 68.68 & 44.04 & 7.61 & 6.59  \\
		&\multirow{2}{*}{$\mathrm{Cos}$}& fixed & 76.46 & 67.96 & 76.62 & 67.54 & 76.33 & 67.39 \\
		&& learn & \bftab{76.79} & \bftab{68.33} & \bftab{76.75} & \bftab{68.18} & \bftab{76.71} & \bftab{68.02}  \\
		\bottomrule
	\end{tabular}
\end{table}

Given the success achieved by our proposed modifications to the vanilla VAE objective, we were curious about the performance of sticking with the original $D_{\mathrm{KL}}\left(q_{\phi}\left(\mathbf{z} | \mathbf{x}, y \right) \Vert p_{\theta}\left(\mathbf{z} | y \right) \right)$ in Eq. \eqref{eq:3-4}. Therefore, we assume $q_{\phi}\left(\mathbf{z} | \mathbf{x}, y \right)$ can be characterized by spherical Gaussians and $p_{\theta}\left(\mathbf{z}, y \right)$ is a MoG prior for the sake of simplicity of the closed-form KL divergence. We report relevant experimental results in Tab. \ref{tab:4-3-3} to demonstrate the superiority of $\mathrm{Cos}\left(\mathbf{z}, \mathbf{c}_y \right)$. Benefited from multiple predictions enabled by our proposed training objective, i.e., the primary predicted outcomes from the recognition component, the auxiliary classifier on latent code $\mathbf{z}$, and the linear separability boosted by $\mathrm{Cos}\left(\mathbf{z}, \mathbf{c}_y \right)$, we decided to report classification metrics on these three types of predictions. We first make comparisons based on the Eq. \eqref{eq:3-4}, which means there are no additional components, such as an auxiliary classifier or $D_{\mathrm{KL}}\left(q_{\eta}\left(y|\mathbf{z}\right) \Vert q_{\phi}\left(y|\mathbf{x}\right) \right)$ involved, as shown in the top part of Tab. \ref{tab:4-3-3}. It should be noted that $D_{\mathrm{KL}}$ with fixed centers for the MoG prior, which are made to be orthonormal, attains performance in all metrics that is only slightly inferior to our proposed TV-PTST because the separability in the latent space is guaranteed by the artificially imposed restriction, implying the benefits of injecting ground-truth information into $D_{\mathrm{KL}}$. Besides, when the prior is made to be learnable\footnote{We used the scheduling scheme in \cite{DBLP:conf/conll/BowmanVVDJB16} to adjust $D_{\mathrm{KL}}$ during the training procedure. Otherwise, the model cannot be trained to convergence.}, as expected, the linear separability in the latent space vanishes, as shown in the Cos column of the second row of Tab. \ref{tab:4-3-3}, although the classification performance of the recognition component remains almost unaffected. Similar findings have already been observed in other work on VAE \cite{DBLP:conf/iclr/0022KSDDSSA17, DBLP:conf/naacl/FuLLGCC19, DBLP:journals/corr/SonderbyRMSW16, DBLP:conf/conll/BowmanVVDJB16} where this problem has been referred to as posterior collapse. Regarding the bottom part of Tab. \ref{tab:4-3-3}, we rerun experiments based on our proposed objective \eqref{eq:3-5}. It is intriguing to observe that $D_{\mathrm{KL}}$ cannot benefit from added components in Eq. \eqref{eq:3-5}, as shown by significant drops in all metrics. By comparing the results in the Z column, it can be seen that latent code obtained by having $D_{\mathrm{KL}}$ instead of $\mathrm{Cos}\left(\mathbf{z}, \mathbf{c}_y \right)$ in Eq. \eqref{eq:3-5} is hardly distinguishable, thereby resulting in extremely low F1 scores and demonstrating the effectiveness of substituting $D_{\mathrm{KL}}$ with $\mathrm{Cos}\left(\mathbf{z}, \mathbf{c}_y \right)$. Besides, regarding the entries in the Cos column, comparing the respective last row of the top and bottom part of Tab. \ref{tab:4-3-3}  can reveal that the added two terms, $\E_{p\left(y| \mathbf{x}\right)}\left[-\log q_{\eta}\left(y | \mathbf{z} \right) \right]$ and $D_{\mathrm{KL}}\left(q_{\eta}\left(y|\mathbf{z}\right) \Vert q_{\phi}\left(y|\mathbf{x}\right) \right)$ in Eq. \eqref{eq:3-5}, can promote linear separability in the latent space significantly, which in turn can improve the performance of the recognition component. Lastly, we want to point out that the three types of predictions Y, Z, Cos obtained from our proposed TV-PTST yield almost identical classification results. Such consistency is crucial to reaching a better solution according to observations made in our experiments.

\begin{table}[!htb]
	\renewcommand{\arraystretch}{1.1}
	\setlength{\tabcolsep}{2.0pt}
	\centering
	\caption{Comparison of GPU memory usage and model parameters of our proposed and state-of-the-art models. N.B., the figures are measured for a single data instance on South Africa dataset.}
	\label{tab:4-3-4}
	\begin{tabular}{cc|c|c|c|c}
		\toprule
		&&\multirow{2}{*}{PSE+L-TAE} & \multirow{2}{*}{PTST} & TV-PTST  & TV-PTST \\
		&& & & (full) & (inf)\\
		\midrule
		\multirow{2}{*}{GPU Mem.(MB)} & Train &  39.14 & \bftab{0.42} & 6.63 & --\\
		& Test & 25.89 & \bftab{0.08} & -- & 0.09\\
		\hline
		\multicolumn{2}{c|}{Model Params.(MB)}& \bftab{0.15} & 0.72 & 1.43 & 0.92 \\
		\bottomrule
	\end{tabular}
\end{table}

Lastly, we compare GPU memory usage and model parameters between our proposed PTST, TV-PTST and the state-of-the-art PSE+L-TAE, as shown in Tab. \ref{tab:4-3-4}. Despite the extreme lightweight of PSE+L-TAE ($0.15$MB), it consumes prohibitively high GPU memory whenever in the train or test phase. On the contrary, the GPU memory consumption of our proposed models is drastically reduced because of the discarded spatial dimensions when treating SITS. We want to point out that TV-PTST (full) consumes more GPU memory in the train phase due to the added decoder. But the recognition component TV-PTST(inf) is the only required one for the test purpose, which means the GPU memory usage still remains at the same level as the purely discriminative classifier PTST. Furthermore, the model capacity is crucial for building powerful deep learning models as demonstrated in a wide range of research areas where model parameters can reach or surpass $100$ or even $1000$ MB, which means PSE+L-TAE would easily encounter computational bottlenecks whenever its model size needs to be enlarged. However, PTST and TV-PTST can be extended to models with large capacity when massive-scale datasets for crop classification are available without the concern for OOM.

\subsection{Comparison with State-of-the-Art}

\begin{table}[!htb]
	\renewcommand{\arraystretch}{1.1}
	\setlength{\tabcolsep}{3.0pt}
	\centering
	\caption{Comparison of our proposed and state-of-the-art models on DENETHOR test dataset.}
	\label{tab:4-3-5}
	\begin{tabular}{l|c|c|c|c}
		\toprule
		& O.A.$\%$ & Precision$\%$ & Recall$\%$ & F1 Score$\%$\\
		\midrule
		PSE+TAE \cite{garnot2020satellite} & 64.95 & -- & -- & 54.50 \\
		PSE+L-TAE \cite{garnot2020lightweight} & 67.25 &-- & -- & 58.12 \\
		PTST & 70.29 & 61.27 & 56.72 & 56.14 \\
		TV-PTST & \bftab{73.78} & \bftab{70.37} & \bftab{65.96} & \bftab{65.50} \\
		\bottomrule
	\end{tabular}
\end{table}

In this section, we report the results of our proposed PTST and TV-PTST on DENETHOR test dataset and make comparisons with state-of-the-art models. As shown in Tab. \ref{tab:4-3-5}, PTST improves O.A. by around \num{3} points but with a lower F1 score (\num{-1.98}) compared to PSE+L-TAE. TV-PTST attains highest performance in all metrics, especially with a significant increase around \num{7} points in F1 score compared to PSE+L-TAE.

\subsection{Semi-Supervised Learning}

\begin{figure}[!htb]
	\centering
	\begin{subfigure}[b]{0.45\textwidth}
		\centering
		\includegraphics[width=\textwidth]{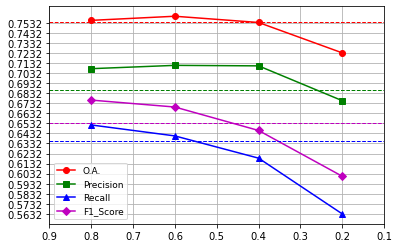}
		\caption{South Africa}
	\end{subfigure}
	\begin{subfigure}[b]{0.45\textwidth}
		\centering
		\includegraphics[width=\textwidth]{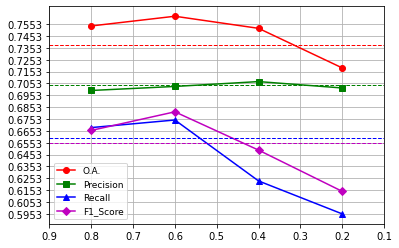}
		\caption{Germany/DENETHOR}
	\end{subfigure}
	\caption{Classification performance of TV-PTST on South Africa and Germany test datasets with partially labelled data. N.B., figures are based on predictions of the recognition component. Please refer to Tab. \ref{tab: e5-1} and \ref{tab: e5-2} in Appendix for detailed results.}
	\label{fig: 4-5-1}
\end{figure}

We have demonstrated that TV-PTST can achieve higher classification performance than its purely discriminative counterpart. Furthermore, we want to showcase its potential in semi-supervised classification, where only a proportion of training samples have corresponding class labels, as we mentioned in Sec. \ref{sec:3-2} that the Eq. \eqref{eq:3-5} is naturally built for dealing with unlabelled data. Besides, it should be noted that we only increased the training epochs from \num{40} to \num{60} for South Africa dataset and from \num{60} to \num{90} for Germany dataset to compensate for the reduction of labelled data and kept other hyper-parameter settings unchanged.

Given that the two public datasets provide full labels, we deliberately remove labels from training instances with various proportions in a way similar to the stratified sampling, resulting in \SI{80}{\percent}, \SI{60}{\percent}, \SI{40}{\percent}, and \SI{20}{\percent} labelled data used for training. As shown in Fig. \ref{fig: 4-5-1}, it is surprising to observe that there is no significant decrease in O.A. and Precision when only \SI{80}{\percent}, \SI{60}{\percent}, and \SI{40}{\percent} labels are used. Actually, the Precision curve on Germany dataset remains flat despite the significant reduction in the proportion of used labels. In addition, we can observe that employing fewer labelled data mainly produces negative effects on Recall, which would also lead to a lower F1 score. Dashed lines in Fig. \ref{fig: 4-5-1} show results of PTST and TV-PTST trained with \SI{100}{\percent} labels on South Africa and Germany datasets, respectively. It can be seen that TV-PTST trained with only \SI{40}{\percent} labelled data can achieve competitive results compared to the purely discriminative classifier PTST on South Africa dataset. Also, TV-PTST trained with only \SI{60}{\percent} labelled data can yield better performance than its fully-supervised trained counterpart. We want to point out that the prolonged training procedure does not benefit PTST or TV-PTST in the fully-supervised setting as it would cause overfitting given the limited number of training samples. Results presented in Fig. \ref{fig: 4-5-1} have shown the potential of our proposed TV-PTST in semi-supervised learning, when at most \SI{60}{\percent} labels are used, which can yield performance that is on par with classifiers trained with \SI{100}{\percent} labelled data.

\section{Conclusion}

In conclude, we have presented a purely discriminative classifier PTST with highly modularized building units and thus easy extensibility. More importantly, we have proposed a series of modifications to the vanilla VAE objective, which have proven to be effective not only in improving classification performance but data efficiency by conducting extensive experiments. Lastly, these two proposed models set new state of the art for crop type mapping using SITS on recently released two datasets. We hope the powerfulness, modularity and transparency of our proposed models can encourage researchers and practitioners in the community to test its potential in a broad array of SITS classification tasks.

\section*{Acknowledgements}

The work is supported by Department for the Economy (DfE) international studentship at Ulster University (UU).

\bibliographystyle{splncs04}	
\bibliography{reference}

\appendix
\renewcommand{\thesection}{\Alph{section}.\arabic{section}}
\setcounter{section}{0}
\begin{appendices}

	\section{Architecture Details}
	
	\begin{figure}[!htb]
		\centering
		\includegraphics[width=0.6\textwidth]{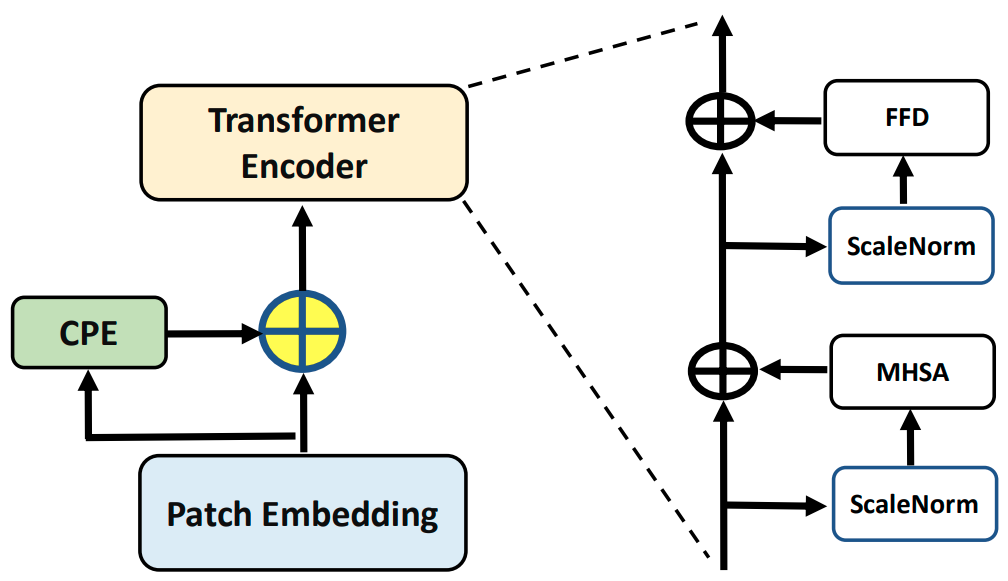}
		\caption{Basic building block for PTST}
		\label{fig: a1-1}
	\end{figure}

	\begin{table}[!htb]
		\renewcommand{\arraystretch}{1.0}
		\setlength{\tabcolsep}{3.0pt}
		\centering
		\caption{Detailed configurations of PTST}
		\label{tab:a1-1}
		\begin{tabular}{c|c|c|c}
			\toprule
			& Output Size & Layer Name & PTST \\
			\midrule
			\hline
			\multirow{3}{*}{Stage 1} & \multirow{3}{*}{$\frac{T}{2}$} & Patch Embedding & $P_1 = 3, S_1=2, C_1=32$  \\ \cline{3-4}
			& & Transformer & \multirow{2}{*}{$\begin{bmatrix} 
					H_1=2 \\
					E_1=1 \\
				\end{bmatrix} \times \left[L_1=1\right] $} \\
			& & Encoder & \\
			\hline
			\multirow{3}{*}{Stage 2} & \multirow{3}{*}{$\frac{T}{4}$} & Patch Embedding & $P_2 = 3, S_2=2, C_2=64$  \\ \cline{3-4}
			& & Transformer & \multirow{2}{*}{$\begin{bmatrix} 
					H_2=4 \\
					E_2=1 \\
				\end{bmatrix} \times \left[L_2=1\right] $} \\
			& & Encoder & \\
			\hline
			\multirow{3}{*}{Stage 3} & \multirow{3}{*}{$\frac{T}{8}$} & Patch Embedding & $P_3 = 3, S_3=2, C_3=128$  \\ \cline{3-4}
			& & Transformer & \multirow{2}{*}{$\begin{bmatrix} 
					H_3=8 \\
					E_3=1 \\
				\end{bmatrix} \times \left[L_3=1\right] $} \\
			& & Encoder & \\
			\hline
			\multirow{3}{*}{Stage 4} & \multirow{3}{*}{$\frac{T}{16}$} & Patch Embedding & $P_4 = 3, S_4=2, C_4=256$  \\ \cline{3-4}
			& & Transformer & \multirow{2}{*}{$\begin{bmatrix} 
					H_4=16 \\
					E_4=1 \\
				\end{bmatrix} \times \left[L_4=1\right] $} \\
			& & Encoder & \\
			\hline
			\bottomrule
		\end{tabular}
	\end{table}

	\begin{table}[!htb]
		\renewcommand{\arraystretch}{1.0}
		\setlength{\tabcolsep}{3.0pt}
		\centering
		\caption{Detailed configurations of the decoder in TV-PTST}
		\label{tab:a1-2}
		\begin{tabular}{c|c|c}
			\toprule
			Output Size & Layer Name & Decoder \\
			\midrule
			\hline
		    \multirow{3}{*}{$T$} & Learnable Position Encoding & $C=128$  \\ \cline{2-3}
			& Transformer & \multirow{2}{*}{$\begin{bmatrix} 
					H=8 \\
					E=1 \\
				\end{bmatrix} \times \left[L=4\right] $} \\
			& Encoder & \\
			
			\hline
			\bottomrule
		\end{tabular}
	\end{table}
\clearpage

\section{Visualization of the Latent Space}

\begin{figure}[!htb]
	\centering
	\begin{subfigure}[b]{0.45\textwidth}
		\centering
		\includegraphics[width=\textwidth]{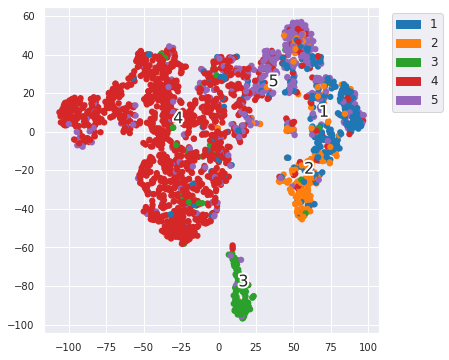}
		\caption{t-SNE South Africa}
	\end{subfigure}
	\begin{subfigure}[b]{0.45\textwidth}
		\centering
		\includegraphics[width=\textwidth]{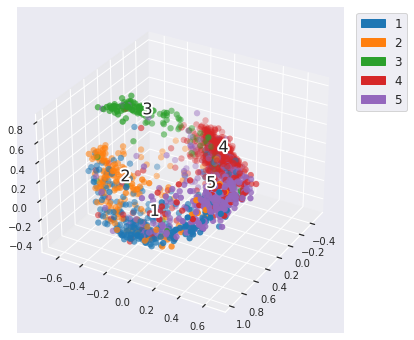}
		\caption{PCA South Africa}
	\end{subfigure}
	\begin{subfigure}[b]{0.45\textwidth}
		\centering
		\includegraphics[width=\textwidth]{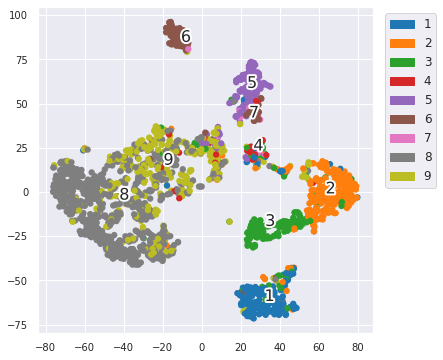}
		\caption{t-SNE Germany}
	\end{subfigure}
	\begin{subfigure}[b]{0.45\textwidth}
		\centering
		\includegraphics[width=\textwidth]{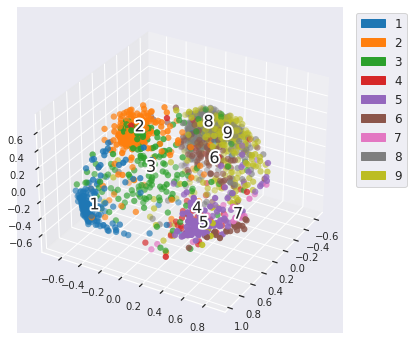}
		\caption{PCA Germany}
	\end{subfigure}
	\caption{Visualization of the latent space on South Africa and Germany test datasets w/ TV-PTST using t-SNE and PCA. Please refer to Fig. \ref{fig: c3-1} for detailed relations between colors and corresponding class labels.}
	\label{fig: b2-1}
\end{figure}

\begin{figure}[!htb]
	\centering
	\begin{subfigure}[b]{0.45\textwidth}
		\centering
		\includegraphics[width=\textwidth]{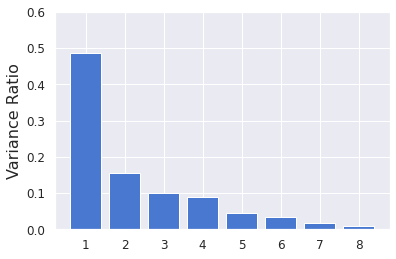}
		\caption{South Africa}
	\end{subfigure}
	\begin{subfigure}[b]{0.45\textwidth}
		\centering
		\includegraphics[width=\textwidth]{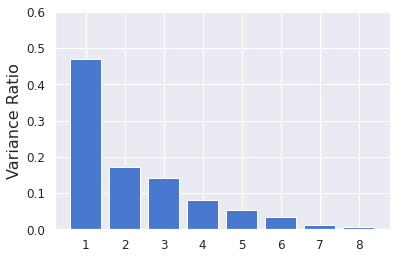}
		\caption{Germany}
	\end{subfigure}
	\caption{Variance ratios of first 8 principal components of latent code $\mathbf{z}$ on South Africa and Germany test datasets w/ TV-PTST.}
	\label{fig: b2-2}
\end{figure}

\begin{figure}[!htb]
	\centering
	\begin{subfigure}[b]{0.75\textwidth}
		\centering
		\includegraphics[width=0.9\textwidth]{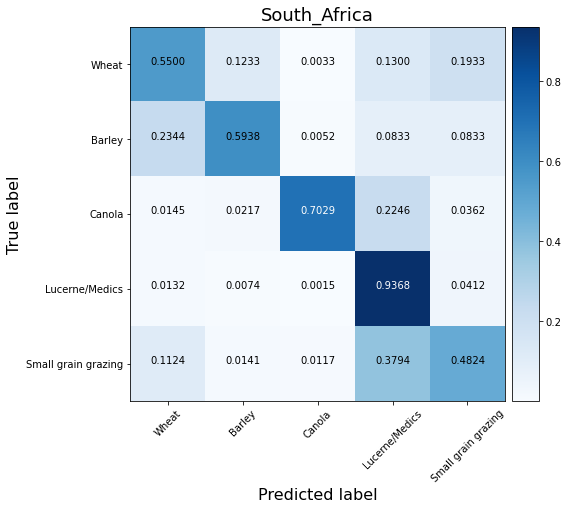}
		\caption{South Africa}
	\end{subfigure}\\
	\begin{subfigure}[b]{0.75\textwidth}
		\centering
		\includegraphics[width=0.9\textwidth]{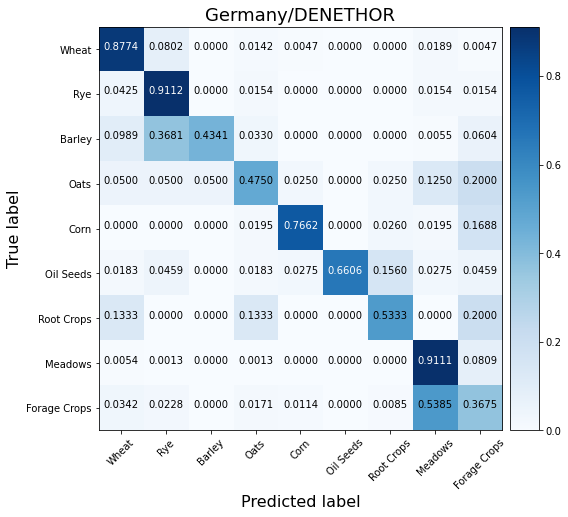}
		\caption{Germany/DENETHOR}
	\end{subfigure}
	\caption{Confusion matrix of TV-PTST}
	\label{fig: d4-1}
\end{figure}

In this section, we visualize the latent space of TV-PTST on South Africa and Germany test datasets, respectively, as illustrated in Fig. \ref{fig: b2-1}. Apart from presenting results obtained using t-SNE \cite{van2008visualizing}, we also show the projection to the first 3 principal axes as we surprisingly found that a few eigenvectors of the latent space with dimensionality of \num{256} where high-dimensional input is mapped to its latent code $\mathbf{z}$ by TV-PTST can capture the majority of variance existing in raw data, as shown in Fig. \ref{fig: b2-2}. Essentially, the first \num{8} principal components account for approximately \SI{95}{\percent} of variance in latent representations of raw data, which explains the high accuracy obtained with cosine similarity scores calculated between latent code $\mathbf{z}$ and class centers. Fig. \ref{fig: b2-1} and \ref{fig: b2-2} well illustrate the increased linear separability as a result of our proposed modifications to the vanilla VAE objective. Additionally, these visualizations can help researchers to identify major failure cases made by TV-PTST, as shown by the consistency between the overlapping areas in latent space and misclassification rates in confusion matrix (Fig. \ref{fig: d4-1}), thereby improving the explainability of our proposed model to a certain extent.

\section{Color Palette Details}
\begin{figure}[!htb]
	\centering
	\begin{subfigure}[b]{0.85\textwidth}
		\centering
		\includegraphics[width=\textwidth]{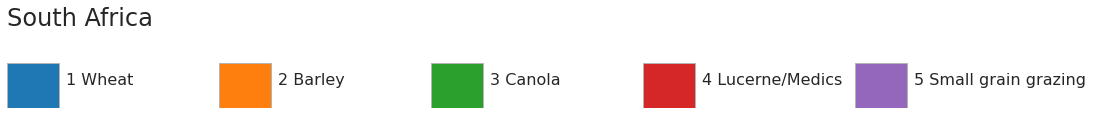}
		\caption{South Africa}
	\end{subfigure}\\
	\begin{subfigure}[b]{0.85\textwidth}
		\centering
		\includegraphics[width=\textwidth]{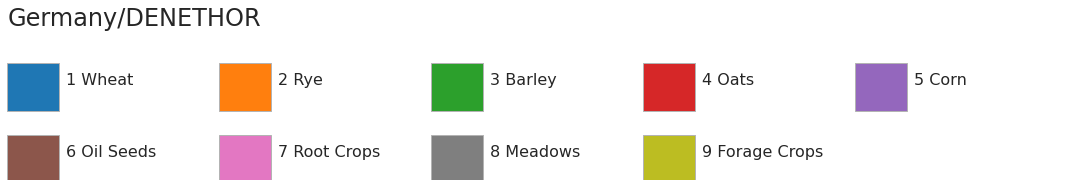}
		\caption{Germany}
	\end{subfigure}
	\caption{Correspondence between colors and class labels used for t-SNE or PCA visualization.}
	\label{fig: c3-1}
\end{figure}
\clearpage

\section{Confusion Matrix}

\clearpage

\section{Detailed Results of Semi-supervised Learning}

\begin{table}[!htb]
	\renewcommand{\arraystretch}{1.1}
	\setlength{\tabcolsep}{3.0pt}
	\centering
	\caption{Detailed classification results of TV-PTST on South Africa test dataset.}
	\label{tab: e5-1}
	\begin{tabular}{c|c|c|c|c|c}
		\toprule
		Labelled Data $\%$ & Prediction Types & O.A. $\%$ & Precision $\%$ & Recall $\%$ & F1 Score $\%$  \\
		\midrule
		\multirow{3}{*}{80} & Y& 75.55 & 70.75 & 65.16 & 67.63 \\
		& Z & \bftab 76.67 & 71.70 & \bftab 65.94 & \bftab 68.45 \\
		& Cos & 76.54 & \bftab 71.81 & 65.48 & 68.22 \\
		\hline
		\multirow{3}{*}{60} & Y& 75.96 & 71.08 & 64.07 & 66.94 \\
		& Z & 76.09 & 71.19 & 64.75 & 67.38 \\
		& Cos & 75.71 & 71.27 & 63.85 & 66.60 \\
		\hline
		\multirow{3}{*}{40} & Y& 75.34 & 71.03 & 61.86 & 64.61 \\
		& Z & 74.14 & 68.89 & 61.56 & 62.96 \\
		& Cos & 74.51 & 68.87 & 62.09 & 63.61 \\
		\hline\multirow{3}{*}{20} & Y& 72.32 & 67.57 & 56.32 & 60.06 \\
		& Z & 70.34 & 63.41 & 54.11 & 55.99 \\
		& Cos & 71.37 & 66.24 & 56.09 & 58.05 \\
		\hline
		\bottomrule
	\end{tabular}
\end{table}

\begin{table}[!htb]
	\renewcommand{\arraystretch}{1.1}
	\setlength{\tabcolsep}{3.0pt}
	\centering
	\caption{Detailed classification results of TV-PTST on Germany/DENETHOR test dataset.}
	\label{tab: e5-2}
	\begin{tabular}{c|c|c|c|c|c}
		\toprule
		Labelled Data $\%$ & Prediction Types & O.A. $\%$ & Precision $\%$ & Recall $\%$ & F1 Score $\%$  \\
		\midrule
		\multirow{3}{*}{80} & Y& 75.38 & 69.94 & 66.81 & 66.57 \\
		& Z & 76.01 & 69.12 & 65.70 & 66.20 \\
		& Cos & 74.70 & 65.16 & 59.99 & 60.71 \\
		\hline
		\multirow{3}{*}{60} & Y& \bftab 76.20 & 70.28 & \bftab 67.45 & \bftab 68.14 \\
		& Z & 74.89 & 66.11 & 63.02 & 63.41 \\
		& Cos & 74.60 & 60.49 & 58.33 & 58.11 \\
		\hline
		\multirow{3}{*}{40} & Y& 75.18 & 70.69 & 62.31 & 64.91 \\
		& Z & 74.84 & 66.85 & 60.23 & 61.96 \\
		& Cos & 74.75 & 60.91 & 58.32 & 58.59 \\
		\hline\multirow{3}{*}{20} & Y& 71.84 & 70.15 & 59.53 & 61.43 \\
		& Z & 71.84 & 71.30 & 58.96 & 61.12 \\
		& Cos & 71.98 & \bftab 75.56 & 56.00 & 58.13 \\
		\hline
		\bottomrule
	\end{tabular}
\end{table}

\end{appendices}
	
\end{document}